\documentclass[10pt,twocolumn,letterpaper]{article}

\usepackage{cvpr}              

\usepackage[dvipsnames]{xcolor}

\definecolor{cvprblue}{rgb}{0.21,0.49,0.74}
\usepackage[pagebackref,breaklinks,colorlinks,citecolor=cvprblue]{hyperref}
\usepackage{graphicx}
\newcommand{\mcircle}[1]{\raisebox{1pt}{\textcircled{\raisebox{-.9pt} {#1}}}}
\newlength\savewidth

\def\paperID{27} 
\def\confName{CVPR}
\def\confYear{2024}

\title{\vspace{-5mm}\textsc{C3DAG}: Controlled 3D Animal Generation using 3D pose guidance}

\author{
Sandeep Mishra$^*$\\
University of Texas at Austin\\
Austin\\
{\tt\small sandy.mishra@utexas.edu}
\and
Oindrila Saha$^*$\\
Univeristy of Massachusetts\\
Amherst\\
{\tt\small osaha@umass.edu}
\and
Alan C. Bovik\\
University of Texas at Austin\\
Austin\\
{\tt\small bovik@ece.utexas.edu}
}

\begin{document}
\twocolumn[{
\renewcommand\twocolumn[1][]{#1}
\maketitle
\begin{center}
\vspace{-5mm}
    \centering
  	\captionsetup{type=figure, width=1\linewidth}
	\includegraphics[width=1\linewidth]{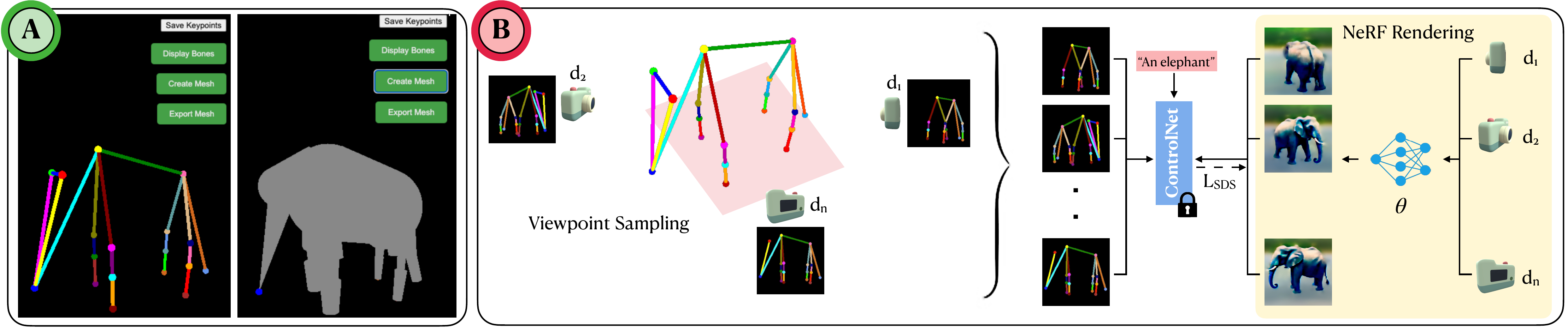}
    \caption{\textbf{\textsc{C3DAG} : A pipeline to generate highly detailed and anatomically accurate 3D Animals.} \mcircle{A} A web-based tool that lets the user modify keypoints in 3D space and generates balloon animals using simple geometric shapes such as spheres, cylinders, and cones with a single button click. The figure depicts the 3D pose and skeleton of an elephant and its generated balloon animal mesh. \mcircle{B} We illustrate our method to create a pose consistent animal by sampling 2D pose control images from a 3D pose based on camera parameters. The sampled 2D poses are used as input to our tetrapod-pose ControlNet along with a text prompt and a NeRF rendered image to backpropagate Score Distillation Sampling (SDS) gradients to the learnable NeRF parameters $\theta$.}
    \label{fig:splash}
\end{center}
}]
\def\thefootnote{*}\footnotetext{These authors contributed equally to this work}
\begin{abstract}
Recent advancements in text-to-3D generation have demonstrated the ability to generate high quality 3D assets. However while generating animals these methods underperform, often portraying inaccurate anatomy and geometry. Towards ameliorating this defect, we present \textsc{C3DAG}, a novel pose-\textbf{C}ontrolled text-to-\textbf{3D} \textbf{A}nimal \textbf{G}eneration framework which generates a high quality 3D animal consistent with a given pose. We also introduce an automatic 3D shape creator tool, that allows dynamic pose generation and modification via a web-based tool, and that generates a 3D balloon animal using simple geometries. A NeRF is then initialized using this 3D shape using depth-controlled SDS. In the next stage, the pre-trained NeRF is fine-tuned using quadruped-pose-controlled SDS. The pipeline that we have developed not only produces geometrically and anatomically consistent results, but also renders highly controlled 3D animals, unlike prior methods which do not allow fine-grained pose control.   
\end{abstract}    
\section{Introduction}
\label{sec:intro}

The past decade has observed dramatic improvements in the field of computer vision, especially on 3D generative tasks. The introduction of denoising diffusion models and implicit representations for 3D reconstruction has dramatically improved the quality of generative art. DreamFusion~\cite{poole2022dreamfusion} showed that knowledge of 2D image generative models can be utilized to guide generation of 3D objects. These concepts alleviate the need for substantial datasets of 3D ground truth, by lifting 2D images and videos into 3D shapes by distillation. With this in mind, we focus on the controlled generation of 3D animals guided by text and pose, with the specific goal of generating anatomically plausible 3D animals. 

3D animal generation has been previously studied, yet current methods still suffer by the creation of problematic anatomical distortions such as multiple heads or limbs, or displaced or missing body parts. Attempts at tackling these problems, such as PerpNeg~\cite{armandpour2023re}, have improved the quality of generation but only in selective cases. Another inherent issue with distillation based methods is the amount of time required to conduct test-time optimization. Although it is not possible to avoid test-time optimization when generating highly detailed 3D assets, efforts to reduce the time required for optimization is an important consideration. 

Another approach to generate 3D animals involves 3D Morphable Models (3DMMs) which are parametric models. While this approach can generate high quality 3D shapes, due to the time-consuming data collection process of 3D scanning, the diversity of generated animals is quite limited. Some methods utilize 2D images and videos to create banks of such models to expand the range of animals they can generate, but out-of-domain generation remains an unsolved problem with this approach. Methods such as MagicPony~\cite{wu2023magicpony}, which use images and videos to learn parametric models, are unable to capture fine details and end up completely omitting tails and other shape details resulting in low quality 3D meshes. 

DreamWaltz~\cite{huang2024dreamwaltz} demonstrates the use of SMPL~\cite{loper2023smpl} based initialization and projected 2D poses to generate highly controlled 3D human-like avatars, while usually reducing inaccuracies in the generated asset. Animals take on a tremendous number of different shapes and body proportions, so starting from an initialization using 3DMM based models like those generated by SMPL is not viable, as they have limited diversity. Towards remediating these limitations, we make the following contributions:

\begin{itemize}
    \item An automatic 3D shape creator tool that generates a naive 3D shape and pose which can be used to initialize a NeRF.
    \item A tetrapod-pose guided ControlNet trained on a diverse dataset including mammals, reptiles, birds and amphibians to provide highly controlled guidance via Score Distilation Sampling (SDS).
    \item We combine (a) and (b) in an efficient two-stage pipeline that is able to generate high quality and anatomically consistent 3D animal assets. Our method, which we call C3DAG, is efficient, and produces a high quality animal much faster than current SOTA methods.
\end{itemize}

\section{Related Work}
\label{sec:related_work}

Generating 3D assets without specific conditions requires understanding the diverse distributions of 3D data. There are two primary strategies: explicit and implicit. Structured representations such as point clouds~\cite{achlioptas2018learning, luo2021diffusion}, voxel grids~\cite{lin2023infinicity, smith2017improved}, and mesh models~\cite{zhang2021sketch2model} are categorized as explicit methods. Implicit techniques generally rely on abstract representations, such as signed distance functions (SDFs)~\cite{chen2019learning, cheng2023sdfusion, mittal2022autosdf}, tri-planes~\cite{chen2023single}, the parameters of multi-layer perceptrons (MLPs)~\cite{erkocc2023hyperdiffusion}, and radiance fields~\cite{lorraine2023att3d}. 

The recent surge in text-to-3D asset generation research has been significantly fueled by the availability of vast datasets of text-image pairs and the success of text-to-image generative models. These innovations have paved the way for the adaptation of pre-trained text-to-image models into the 3D domain, primarily those that are CLIP-guided, or are 2D diffusion-guided. CLIP-guided methods such as~\cite{mohammad2022clip, jain2022zero} leverage cross-modal matching models like CLIP~\cite{radford2021learning} for text-to-3D conversion, whereas diffusion-guided strategies utilize text-to-image diffusion models~\cite{zhu2023hifa, chen2023fantasia3d}, like Imagen~\cite{saharia2022photorealistic} and Stable Diffusion~\cite{rombach2022high}, to generate 3D assets from textual descriptions. The latter has been noted to deliver superior text-to-3D generation performance, employing techniques such as Score Distillation Sampling (SDS)~\cite{poole2022dreamfusion}, which refines noise in images captured from NeRFs~\cite{mildenhall2021nerf}, and Score-Jacobian-Chaining~\cite{wang2023score}, which aggregates image gradients into 3D asset gradients. 

The field of 3D animal generation has made significant progress thanks to groundbreaking studies that have provided new methods and insights for modeling the complex structures and movements of animals in 3D. SMAL~\cite{Zuffi:CVPR:2017} introduced a method to fit a parametric 3D shape model, derived from 3D scans, to animal images using 2D keypoints and segmentation masks, with extensions to multi-view images~\cite{zuffi2018lions}. Subsequent efforts, such as LASSIE~\cite{yao2022lassie}, have focused on deriving 3D shapes directly from smaller image collections by identifying self-supervised semantic correspondences to discover 3D parts. Despite these advances, the limited diversity of currently available 3D assets confines most research~\cite{jakab2023farm3d, wu2023dove, wu2023magicpony} to the production of 3D models that are both class-specific and small scale.

\section{Method}
\label{sec:method}

We propose \textsc{C3DAG}: Controlled 3D Animal Generation, which is an efficient two-stage approach to generate high-quality 3D assets visually representative of posed tetrapods, including birds, animals and reptiles. The first stage initializes a NeRF to a generic balloon animal created using our automatic 3D shape creator tool. The second stage refines the initialization by using occlusion aware quadruped-3D-pose consistent Score Distillation Sampling (SDS) to generate high quality 3D animal assets. Unlike DreamWaltz \cite{huang2024dreamwaltz}, we avoid using 3DMM models, since they are always limited to a fixed set of animals, thus enabling \textsc{C3DAG} to generate a more diverse set of fauna including quadruped land animals, reptiles and birds. We also introduce control scale and guidance scale annealing, ensuring that the 3D optimization is able to fully utilize the diversity and quality of Stable Diffusion. See Supplementary Sec.~\ref{sec:preliminaries} for more details about the prelimanary concepts used in this article. Hereafter we describe each aspect in detail.

\subsection{2D Pose guided ControlNet}
\label{sec:2D Pose guided ControlNet}
Following the human-pose ControlNet introduced in the original paper, we train a ControlNet that can produce various animal images including quadrupeds, reptiles and birds using the same set of 18 keypoints. We introduce data augmentations such as random rotations, translations, and scaling, so that the model is robust against various factors such as occluded poses and different scales of animals. This is crucial in the 3D generation stage as the randomly sampled camera views of a 3D skeleton will often include heavy occlusions. We provide more details in the section on Implementation Details.


\begin{figure*}[t]
    \centering
    \vspace{-3mm}
    \includegraphics[width=1\textwidth]{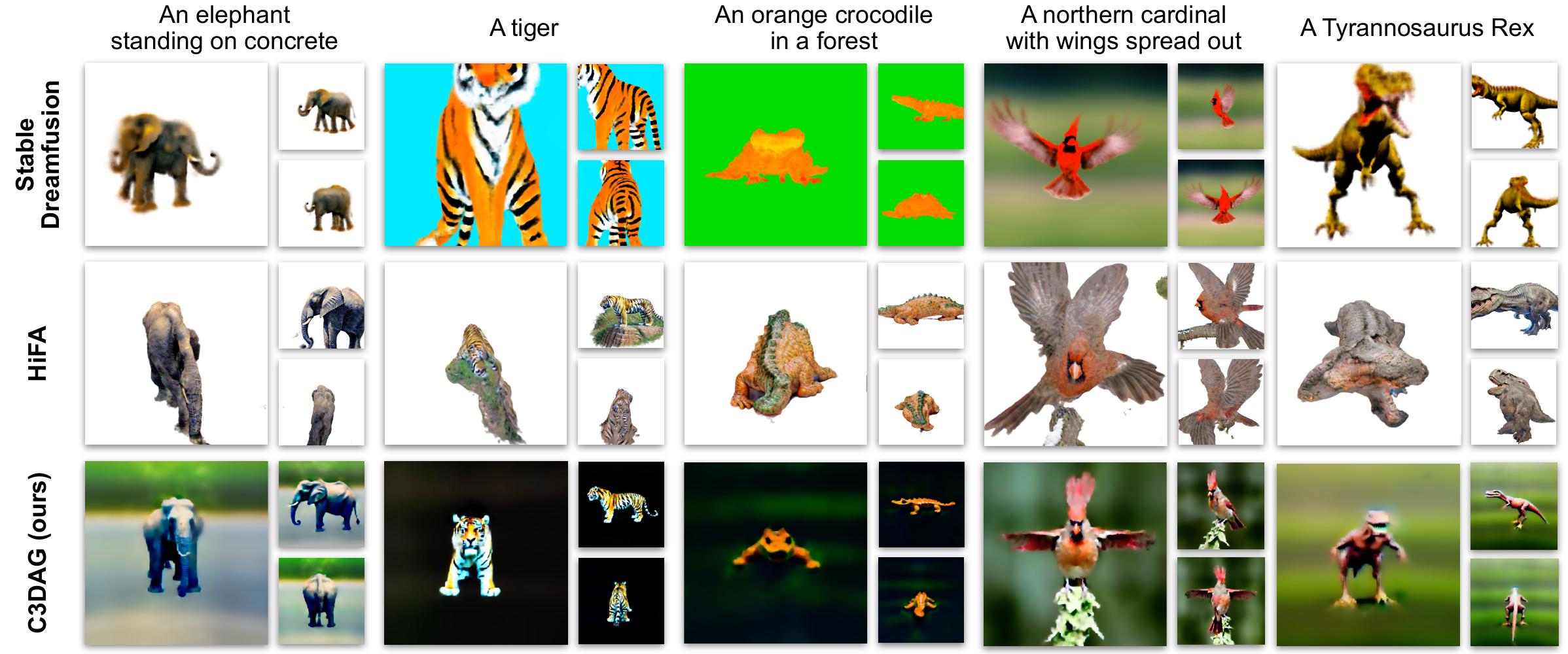}
    \captionsetup{justification=justified}
    \caption{\textbf{Comparison with text-to-3D generation guided by 2D diffusion.} Given the prompt as shown, we qualitatively compare with the open-source version of Dreamfusion and HiFA using their default settings. For both of these we append ``, full body" to the prompt. It is clearly observed that both prior work suffer from various inconsistencies producing multiple heads or limbs. Stable Dreamfusion usually produces lower details in textures. HiFA produces high-quality textures but almost always produces anatomically incorrect animals. We provide more results in supplementary.}
    \vspace{-5mm}
    \label{fig:comparetextto3d}
\end{figure*}

\subsection{NeRF initialization}
\label{sec:NeRF initialization}
To conduct NeRF pre-training using depth ControlNet, DreamWaltz uses the neutral SMPL mesh. However initialization using a single generic model is ineffective for generating tetrapods, as they vary vastly in shape and size. Existing parametric models such as SMAL~\cite{Zuffi:CVPR:2017} also cannot be used, since they represent only a small set of animals that does not include any birds or reptiles. Another parametric model called 3DFauna~\cite{li2024learning}, has wider diversity, but also fails to represent elephants, and appendages like tails in certain animals, as evidenced in the results in their original paper and website. To remediate these issues, we instead resort to an automatic method of generating an initial 3D mesh given a 3D skeleton, as described in Sec.~\ref{sec:Generating 3D pose and shape}. The effect of using this pre-training is shown in supplementary.



\subsection{Generating 3D pose and shape}
\label{sec:Generating 3D pose and shape}
We employ a set of 18 3D keypoints and 18 bone connections to represent a generic body pose, as visualized in Fig.~\ref{fig:splash} \mcircle{A}. To enable the modification of the 3D pose we created a simple THREE.js web tool that provides a clean UI to move the 3D keypoints around. The same tool also allows the user to initialize the 3D shape using spherical, cylindrical and conical meshes to form the head, limbs, tails, and nose, respectively. The parameters of the constructive components are tunable, enabling the creation of balloon animals of different shapes and sizes. The combined components are then re-meshed into a single mesh, which is then used in the pre-training stage using SDS, with depth ControlNet as the guidance.

\subsection{3D Aware Score Distillation Sampling}
\label{sec:3D aware Score Distillation Sampling}
Given the input 3D keypoints we project 2D poses onto the camera coordinate system, then use the projections as control images when generating pose-based-guidance signals from our trained Tetrapod-Pose guided ControlNet. Since the 3D pose is fixed, this makes it possible to produce 3D consistent animal views, ensuring a high quality reconstruction. We illustrate this in Fig.~\ref{fig:splash} \mcircle{B}. It is important to note that \textsc{C3DAG} uses no 3D ground-truth data unlike MVDream~\cite{shi2023mvdream} or Zero-1-2-3~\cite{liu2023zero}, instead purely lifting 2D poses to 3D, guided only by training on 2D image datasets. We show that without pose control, other 2D diffusion SDS guided methods fail to maintain geometric and anatomic consistency (Fig. \ref{fig:comparetextto3d}). However, since ControlNet is trained using a limited number of images from datasets that contain annotated poses, exploiting the quality and diversity learned by stable diffusion is important. Animal pose datasets inherently consider occlusion, as only visible keypoints are annotated. In light of this, we deployed a more generic variation of occlusion culling than that in DreamWaltz, to control the visibility of each individual part of the head (left eye, right eye, and nose) depending on the view-description. We utilized randomly acquired camera parameters to determine the view description, including: front, left-side, back, right-side, top and bottom.


\begin{figure}[t]
    \centering
    \includegraphics[width=1\columnwidth]{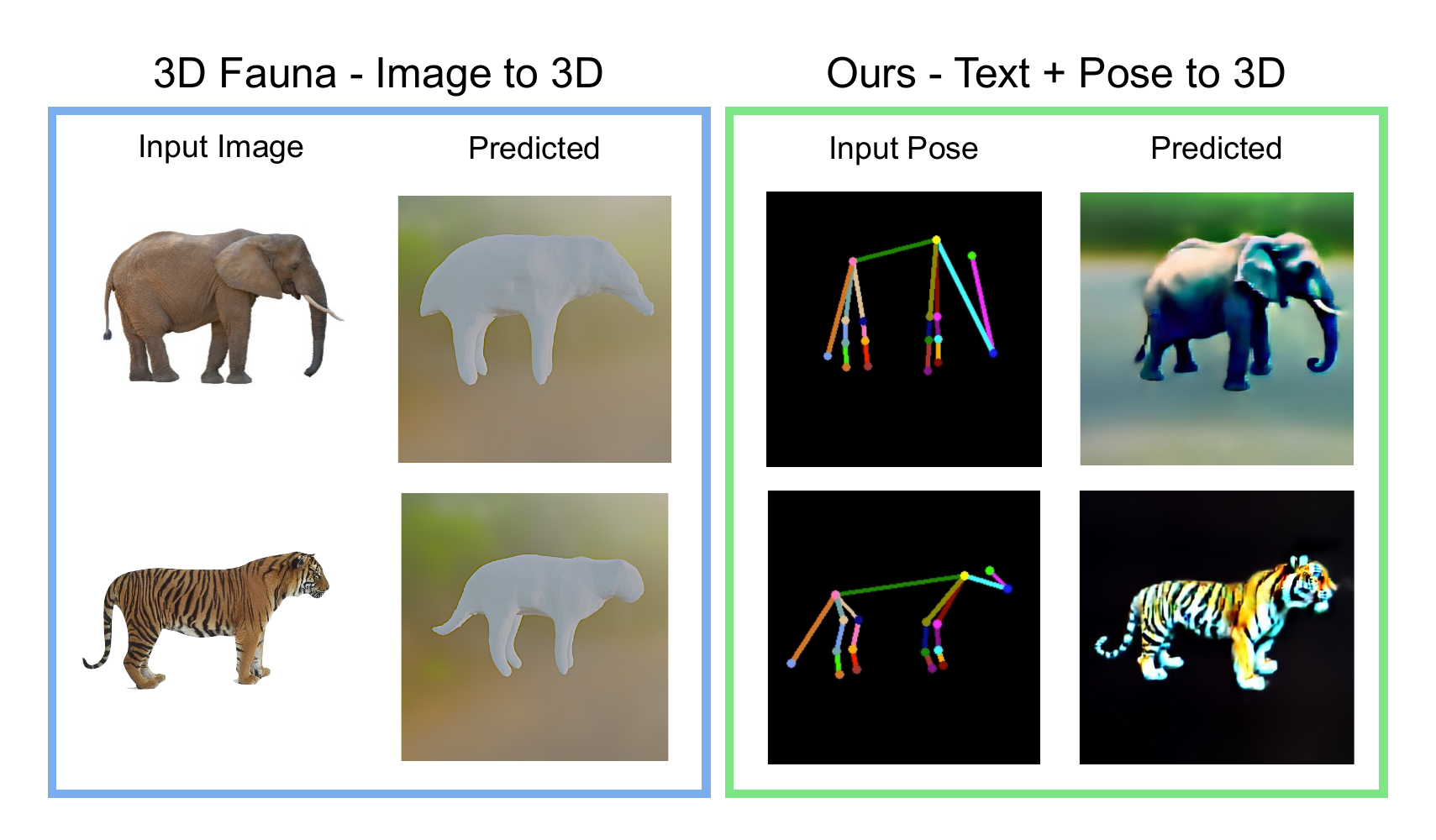}
    \captionsetup{justification=justified}
    \caption{\textbf{Comparison with parametric model based method.} Given the input image 3DFauna fails to capture high-frequency details and follow the input image (see tail), whereas our method produces a highly detailed animal given input pose and text, which closely follows the input pose control.}
    \vspace{-5mm}
    \label{fig:ours_vs_3Dfauna}
\end{figure}

\section{Implementation details}
\textbf{ControlNet Training:} We used annotated poses from the AwA-pose~\cite{banik2021novel} and Animal Kingdom~\cite{ng2022animal} datasets to train ControlNet in a similar way as the original, which uses stable diffusion version 1.5. AwA-pose consists of 10k annotated images covering 35 quadruped animal classes, while Animal Kingdom provides 33k annotated images spanning 850 species, including mammals, reptiles, birds, amphibians, fishes and insects. From a combined set of 43k samples, we carefully selected a subset including only mammals, reptiles, birds, and amphibians. We also eliminated any sample having less than 30\% of its keypoints annotated. The curated dataset consists of 13k annotated samples. To increase diversity in learning, and to improve test-time generation at any scale and transformations, we used a combination of data augmentation strategies consisting of random rotations, translations, and scaling while training. The model was trained over 229k iterations with a batch size of 12, a constant learning rate of $1e^{-5}$, on a single Nvidia RTX 6000. The model converged after around 120k iterations and would not overfit even up to 200k iterations, owing in part to the augmentation strategy. 

\textbf{3D Pose and Shape generation:} We used the following 18 keypoints to represent every quadruped: left eye, right eye, nose, neck end, 4 $\times$ thighs, 4 $\times$ knees, 4 $\times$ paws, back end, and tail end. For the upper limbs of birds, i.e. wings, their front - {thighs, knees, and paws} are defined in accordance with how their upper limbs move. We began with an initial pose of a tiger from SMAL, and modify its keypoints using the balloon animal creator tool. An example of this process and the tool UI is depicted in Fig. \ref{fig:splash}\mcircle{A}. After appropriate modification of the pose the user can press the button to \textit{create mesh} around bones. This button press invokes calls to various functions defined to create each body part, based on their natural appearances using simple mesh components such as ellipses, cylinders, and cones. The combined mesh and the corresponding keypoints can be downloaded by another button click. 

\textbf{Mesh depth guided NeRF initialization}: The mesh downloaded in the previous step was used to provide depth maps to the pre-trained depth guided ControlNet, which produces the gradient loss by SDS, which in turn is used to pre-train the NeRF. The pre-training helps achieve a reasonable initial state for the NeRF weights, which can then be refined in the final pose-guided training stage. The diffusion model was pre-trained for 10,000 iterations using the Adam optimizer with a learning rate of $1e-3$ and a batch size of 1. During training, the camera positions were randomly sampled in spherical coordinates, where the radius, azimuthal angle, and polar angle of camera position were sampled from [1.0, 2.0], [0, 360], and [60, 120].

\textbf{Pose-guided SDS for NeRF fine-tuning:} Finally, we fine-tune the NeRF using the pre-trained ControlNet to provide 2D pose guidance to SDS. The gradients computed using the noise residual from SDS was weighted in a similar manner as DreamFusion, where $w(t) = \sigma_{t}^{2}$ and $t$ was annealed using $t = t_{max} - (t_{max} - t_{min})\sqrt{\frac{iter}{total\_iters}}$. We set $t_{max}$ to be 0.98, $t_{min}$ to be $0.4$. Similar to the previous stage, we trained the model over $total\_iters = 10,000$ using the same settings for the optimizer. Using cosine annealing, we reduced the control scale from an initial value of 1 to a final value of 0.25, while updating guidance scale as $\omega = \omega_{init}\cdot(1+\frac{iter}{total\_iters})$, where $\omega_{init} = 50$. These settings helped reduce the impact of ControlNet gradually over the training process, while improving quality by gradually increasing $\omega$ of classifier-free guidance. The camera positions were randomly sampled as in stage 1, as were the radius, azimuthal angle, and polar angle of the camera. The 3D avatar representation renders “latent images” in the latent space of $\mathbb{R}^{64\times64\times4}$ following LatentNeRF~\cite{metzer2023latent}, where the “latent images” can be decoded into RGB images of $\mathbb{R}^{512\times512\times4}$ using the VAE decoder of Stable Diffusion~\cite{rombach2022high}.

\section{Discussion and Conclusion}
\label{sec:discussion}
We showed our method performs consistently well (Fig~\ref{fig:comparetextto3d}) while Stable DreamFusion and HiFA usually produce physically implausible or low-quality assets.  It is important to note that HiFA, which is a current SOTA model, requires about 7 hours on a 80GB A100 GPU, while ours needs only about 20 minutes on the same setting. Fig~\ref{fig:ours_vs_3Dfauna} shows that our generated animal contains much more detail than the one produced by 3DFauna~\cite{li2024learning}. Our method currently requires some human effort to re-position 3D keypoints in our balloon animal creator tool. It is a direction of future work to fully automate this process. To summarize, we present \textsc{C3DAG} -- an efficient method to generate anatomically and geometrically consistent 3D animals. We contribute 1) a 2D Tetrapod-pose ControlNet trained on mammals, amphibians, reptiles and birds, 2) a tool to create 3D balloon animals automatically given 3D keypoints, and 3) a method to pipeline 1) and 2) to create high quality 3D animals.

{
    \small
    \bibliographystyle{ieeenat_fullname}
    \bibliography{main}
}

\clearpage
\setcounter{page}{1}
\maketitlesupplementary


\section{Preliminaries}
\label{sec:preliminaries}
Recent works have shown impressive results in 2D and 3D content generation using a combination of the following:

\begin{itemize}
    \item \textbf{Diffusion Model:} \\ 
    Generative pre-training on extensive image-text datasets using denoising diffusion models serves as effective priors for text-to-3D generation. These models undergo two primary processes:

    \textit{Forward Process:} This involves incrementally adding noise to the data \(x \sim p(x)\), transforming it towards a Gaussian distribution over \(T\) steps. The noise-augmented data at step \(t\) is described by:
    \[ z_t = \sqrt{\overline{\alpha}_t} x + \sqrt{1 - \overline{\alpha}_t} \epsilon, \quad \epsilon \sim \mathcal{N}(0, I), \]
    where \(\overline{\alpha}_t = \prod_{s=1}^{t} \alpha_s\) and \(\alpha_t \in (0, 1)\) as part of the predefined noising schedule.
    
    \textit{Backward Process:} In this phase, the model learns to reverse the noise addition, aiming to reconstruct the original data from its noised version. The models learns to estimate the noise by minimizing
    \[ L_t = \mathbb{E}_{x, \epsilon \sim \mathcal{N}(0,I)}\left[ \|\epsilon - \epsilon_\phi(z_t, t)\|^2 \right].\]
    
    This dual process enables the diffusion model to accurately estimate and reconstruct \(x\) from noisy observations, effectively leveraging clean data from noised counterparts.

    \item \textbf{Neural Radiance Fields (NeRF):} \\
    Widely embraced for text-to-3D generation tasks, NeRFs utilize a trainable Multilayer Perceptron (MLP) parameterized by \(\theta\) as their core 3D representation. When rendering, a set of rays \(r(k) = o + kd\) are sampled, where \(o\) represents the camera position and \(d\) the direction, calculated on a per-pixel basis. The MLP processes each ray \(r(k)\), outputting the density \(\tau\) and color \(c\) for each sampled point. The final color \(C\) of a pixel is then determined by approximating the volume rendering integral using numerical quadrature:\[ \hat{C}(r) = \sum_{i=1}^{N_c} \Omega_i \cdot (1 - \exp(-\tau_i\delta_i))c_i, \] where \(N_c\) is the number of points sampled along a ray, \(\Omega_i = \exp\left(-\sum_{j=1}^{i-1} \tau_j \delta_j\right)\) is the accumulated transmittance, and $\delta_i$ is the distance between adjacent sample points.

    \item \textbf{Score Distillation Sampling (SDS):} \\
    Introduced by DreamFusion and further utilized in various studies, SDS is a method designed to transfer the knowledge from a pre-trained diffusion model \(\epsilon_\phi\) into a differentiable 3D representation, enhancing a NeRF model's parameters \(\theta\). The rendered output \(x\) from a NeRF model can be obtained via \(x = g(\theta)\), where \(g\) signifies a differentiable rendering function. The core of SDS lies in computing the gradients of the NeRF parameters \(\theta\), formulated as:
    \[ \nabla_\theta L_{\text{SDS}}(\phi, x) = \mathbb{E}_{t,\epsilon}\left[ w(t) \left(\epsilon_\phi(x_t; y, t) - \epsilon\right) \frac{\partial z_t}{\partial x} \frac{\partial x}{\partial \theta} \right], \] where \(w(t)\) is a timestep-dependent weighting function and \(y\) is the input text prompt. 
    \begin{figure}[t!]
    \centering
    \includegraphics[width=1\columnwidth]{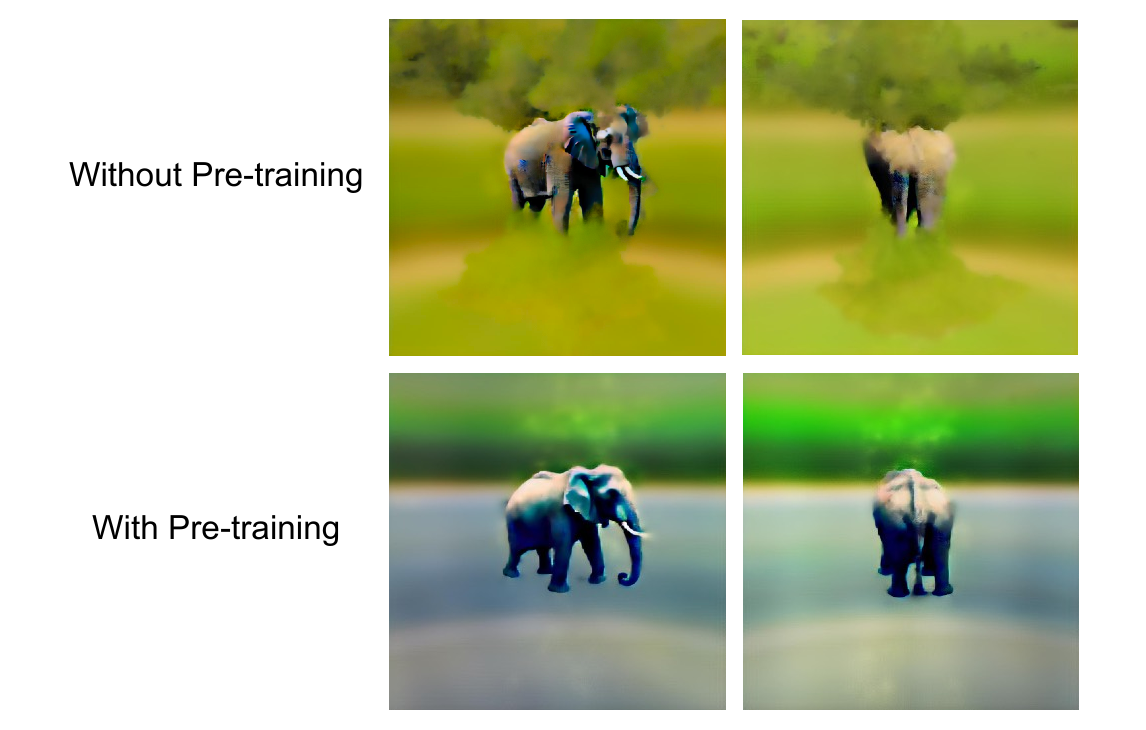}
    \captionsetup{justification=justified}
    \caption{\textbf{Effect of NeRF pre-training using shape generated by the automatic 3D shape creator tool.} The top row shows the final generated 3D without any initial pre-training stage. This has visible shortcomings in form of body occlusion as well as extra volume in the foreground which does not belong to the body of the animal. In contrast, fine-tuning after a pre-training stage generates a clean result.}
    \label{fig:worwopretrain}
    \end{figure}

    \item \textbf{ControlNet guided SDS:} \\
    A novel approach that extends the principles of Score Distillation Sampling (SDS) to include conditioning on additional image-based information, facilitating enhanced control when generating 3D models. Like SDS, ControlNet~\cite{zhang2023adding} leverages a pre-trained diffusion model \(\epsilon_\phi\) but also introduces a conditioning image \(c\), which could be a skeleton/pose map, depth map, normal map, or a combination thereof, to guide the generation process more precisely. For a model parameterized by \(\theta\), and its output \(x\), ControlNet refines the gradient calculation of the parameters \(\theta\) by incorporating \(c\), as shown in: \[ \nabla_\theta L_{\text{SDS}}(\phi, x) = \mathbb{E}_{t,\epsilon}\left[ w(t) \left(\epsilon_\phi(x_t; y, t, \textcolor{red}{c}) - \epsilon\right) \frac{\partial z_t}{\partial x} \frac{\partial x}{\partial \theta} \right], \] where \(w(t)\) is a weighting function dependent on the timestep \(t\), \(y\) denotes the given text prompt, and \(c\) is the conditioning image that significantly influences the generation process. This approach benefits by ControlNet's innovative method of integrating additional contextual cues into the generative framework, thereby enriching the model's capacity to compute detailed and accurate 3D reconstructions using complex conditioning information.

\end{itemize}

\section{Impact of NeRF Pre-training}
\label{sec:NeRFPre-trainingImpact}

Fig.~\ref{fig:worwopretrain} shows the impact of the NeRF pre-training stage using depth ControlNet as described in Sec~\ref{sec:NeRF initialization} using the 3D balloon animal generated by the automatic 3D shape creator tool. This shows that using the pre-training stage results in visibly cleaner results without any body occlusions.


\end{document}